\pdfoutput=1

\documentclass[11pt]{article}

\usepackage{EMNLP2023}

\usepackage{times}
\usepackage{latexsym}
\usepackage{caption}
\usepackage{subcaption}
\usepackage{tabularx}
\usepackage{graphicx}
\usepackage{amsmath} 
\usepackage{commath}

\usepackage[T1]{fontenc}

\usepackage[utf8]{inputenc}

\usepackage{microtype}

\usepackage{inconsolata}

\title{Filling in the Gaps: \protect\\
Efficient Event Coreference Resolution using Graph Autoencoder Networks}

\author{Loic De Langhe, Orphée De Clercq, Veronique Hoste \\
         LT3, Language and Translation Technology Team, Ghent University, Belgium \\ Groot-Brittanniëlaan 45, 9000 Ghent, Belgium \\ \texttt{firstname.lastname@ugent.be}}

\begin{document}
\maketitle
\begin{abstract}
We introduce a novel and efficient method for Event Coreference Resolution (ECR) applied to a lower-resourced language domain. By framing  ECR as a graph reconstruction task, we are able to combine deep semantic embeddings with structural coreference chain knowledge to create a parameter-efficient family of Graph Autoencoder models (GAE). Our method significantly outperforms classical mention-pair methods on a large Dutch event coreference corpus in terms of overall score, efficiency and training speed. Additionally, we show that our models are consistently able to classify more difficult coreference links and are far more robust in low-data settings when compared to transformer-based mention-pair coreference algorithms. \end{abstract}

\section{Introduction}\label{sec:intro}

Event coreference resolution (ECR) is a discourse-centered NLP task in which the goal is to determine whether or not two textual events refer to the same real-life or fictional event. While this is a fairly easy task for human readers, it is far more complicated for AI algorithms, which often do not have access to the extra-linguistic knowledge or discourse structure overview that is required to successfully connect these events. Nonetheless ECR, especially when considering cross-documents settings, holds interesting potential for a large variety of practical NLP applications such as summarization \cite{liu2019hierarchical}, information extraction \cite{humphreys_event_1997} and content-based news recommendation \cite{8612730}.

However, despite the many potential avenues for ECR, the task remains highly understudied for comparatively lower-resourced languages. Furthermore, in spite of significant strides made since the advent of transformer-based coreference systems, a growing number of studies has questioned the effectiveness of such models. It has been suggested that classification decisions are still primarily based on the surface-level lexical similarity between the textual spans of event mentions \cite{ahmed20232, electronics12040850}, while this is far from the only aspect that should be considered in the classification decision. Concretely, in many models coreferential links are assigned between similar mentions even when they are not coreferent, leading to a significant number of false positive classifications, such as between Examples \ref{ex:problem1} and \ref{ex:problem2}.

\begin{enumerate}
\setcounter{enumi}{0}
  \item \label{ex:problem1} 
  The French president Macron met with the American president for the first time today
  \item \label{ex:problem2} 
  French President Sarkozy met the American president
\end{enumerate}

We believe that the fundamental problem with this method stems from the fact that in most cases events are only compared in a pairwise manner and not as part of a larger coreference chain. The evidence that transformer-based coreference resolution is primarily based on superficial similarity leads us to believe that the current pairwise classification paradigm for transformer-based event coreference is highly inefficient, especially for studies in lower-resourced languages where the state of the art still often relies on the costly process of fine-tuning large monolingual BERT-like models \cite{de2022investigating}. 

In this paper we aim to both address the lack of studies in comparatively lower-resourced languages, as well as the more fundamental concerns w.r.t. the task outlined above. We frame ECR as a graph reconstruction task and introduce a family of graph autoencoder models which consistently outperforms the traditional transformer-based methods on a large Dutch ECR corpus, both in terms of accuracy and efficiency. Additionally, we introduce a language-agnostic model variant which disregards the use of semantic features entirely and even outperforms transformer-based classification in some situations. Quantitative analysis reveals that the lightweight autoencoder models can consistently classify more difficult mentions (cfr. Examples 1 and 2) and are far more robust in low-data settings compared to traditional mention-pair algorithms.

\section{Related Work}

\subsection{Event Coreference Resolution}
 
 The primary paradigm for event coreference resolution takes the form of a binary mention-pair approach. This method generates all possible event pairs and reduces the classification to a binary decision (coreferent or not) between each event pair. A large variety of classical machine learning algorithms has been tested using the mention-pair paradigm such as decision trees \cite{cybulska_translating_2015}, support vector machines \cite{chen_event_2015} and standard deep neural networks \cite{nguyen2016new}. 
 
 More recent work has focused on the use of LLMs and transformer encoders \cite{cattan2021cross, cattan2021realistic}, with span-based architectures attaining the best overall results \cite{joshi2020spanbert, lu2021conundrums}. It has to be noted that mention-pair approaches relying on LLMs suffer most from the limitations discussed in Section \ref{sec:intro}. In an effort to mitigate these issues some studies have sought to move away from the pairwise computation of coreference by modelling coreference chains as graphs instead. These methods' primary goal is to create a structurally-informed representation of the coreference chains by integrating the overall document \cite{fan2022enhancing, tran2021exploiting} or discourse \cite{huang2022incorporating} structure. Other graph-based methods have focused on commonsense reasoning \cite{wu2022cross}. 

 Research for comparatively lower-resourced languages has generally followed the paradigms and methods described above and has focused on languages such as Chinese \cite{mitamura_event_2015}, Arabic \cite{nist_ace_2005} and Dutch \cite{minard_meantime_2016}.
 
\subsection{Graph Autoencoders}\label{sec:relgae}

Graph Autoencoder models were introduced by \citet{kipf2016variational} as an efficient method for graph reconstruction tasks. The original paper introduces both variational graph autoencoders (VGAE) and non-probabilistic graph autoencoders (GAE) networks. The models are parameterized by a 2-layer graph-convolutional network (GCN) \cite{kipf2016semi} encoder and a generative inner-product decoder between the latent variables. While initially conceived as lightweight models for citation network prediction tasks, both the VGAE and GAE have been successfully applied to a wide variety of applications such as molecule design \cite{liu2018constrained}, social network relational learning \cite{yang2020relation} and 3D scene generation \cite{chattopadhyay2023learning}. Despite their apparent potential for effectively processing large amounts of graph-structured data, application within the field of NLP has been limited to a number of studies in unsupervised relational learning \cite{li2020r}. 

\section{Experiments}

\begin{table*}[t]
\centering
\scalebox{0.75}{
\begin{tabular}{lccccc}
\textbf{Model}      & \textbf{CONLL F1} & \textbf{Training Runtime (s)} & \textbf{Inference Runtime (s)} & \textbf{Trainable Parameters} & \textbf{Disk Space (MB)} \\ \hline
MP RobBERTje        &       0.767
         &          7962              &             16.31            &        74M          & 297               \\ 
MP BERTje$_{ADPT}$     &      0.780          &      12 206                  &        20.61                 &     0.9M       &     3.5                \\
MP BERTje           &      0.799          &       9737                 &       21.78                  &          110M      &      426           \\ \hline
GAE NoFeatures          &         0.832 ± 0.008
       &           1006             &         0.134              &            825856  &  3.2 
                     \\
GAE BERTje$_{768}$    &     0.835 ± 0.010
           &       975                 &        0.263                &             51200       &     0.204        \\
GAE BERTje$_{3072}$   &      \textbf{0.852 ± 0.006}
          &        1055                &            0.294             &           198656       &        0.780       \\
GAE RobBERT$_{768}$    &       0.838  ± 0.004       &     1006                   &           0.273              &              51200     &     0.204         \\
GAE RobBERT$_{3072}$  &        0.841  ± 0.007      &       1204                 &        0.292                 &          198656      &      0.780          \\
GAE SBERT           &        0.801 ± 0.002
        &          982              &             0.291            &               51200     &    0.204         \\ \hline
VGAE NoFeatures         &        0.824 ± 0.009
        &          1053              &             0.139            &             827904 &  3.2 
                    \\
VGAE BERTje$_{768}$    &     0.822  ± 0.011         &      1233                  &        0.282                 &         53248         &   0.212            \\
VGAE BERTje$_{3072}$  &       0.842 ± 0.009        &       1146                 &      0.324                   &        200704      &    0.788              \\
VGAE RobBERT$_{768}$   &     0.828 ± 0.0021          &     1141                   &       0.288                  &            53248     &    0.212            \\
VGAE RobBERT$_{3072}$ &    0.831  ± 0.004         &         1209               &       0.301                  &           200704    &      0.788            \\
VGAE SBERT          &       0.773 ± 0.012
         &              1185          &            0.295             &           53248     &     0.212           
\end{tabular}}
\caption{Results for the cross-document event coreference task. We report the average CONLL score and standard deviation over 3 training runs with different random seed initialization for the GCN weight matrices (GAE/VAE) and classification heads (Mention-Pair models). Inference runtime is reported for the entire test set.}
\end{table*}

\subsection{Data}
Our data consists of the Dutch ENCORE corpus \cite{de2022constructing}, which in its totality consists of 12,875 annotated events spread over 1,015 documents that were sourced from a collection of Dutch (Flemish) newspaper articles. Coreferential relations between events were annotated at the within-document and cross-document level. 
\subsection{Experimental Setup}

\subsubsection{Baseline Coreference Model}
Our baseline model consists of the Dutch monolingual BERTje model \cite{de2019bertje} fine-tuned for cross-document ECR. First, each possible event pair in the data is encoded by concatenating the two events and by subsequently feeding these to the BERTje encoder. We use the token representation of the classification token \textit{[CLS]} as the aggregate embedding of each event pair, which is subsequently passed to a softmax-activated classification function. Finally, the results of the text pair classification are passed through a standard agglomerative clustering algorithm \cite{kenyon-dean_resolving_2018, barhom2019revisiting} in order to obtain output in the form of coreference chains. 

We also train two parameter-efficient versions of this baseline model using the distilled Dutch Language model RobBERTje \cite{delobelle2022robbertje} and a standard BERTje model trained with bottleneck adapters \cite{pfeiffer2020adapterhub}.

\subsubsection{Graph Autoencoder Model}

We make the assumption that a coreference chain can be represented by an undirected, unweighted graph $\mathcal{G}$ = (\textit{V}, \textit{E}) with $\abs{V}$ nodes, where each node represents an event and each edge $\textit{e} \in \textit{E}$ between two nodes denotes a coreferential link between those events. We frame ECR as a graph reconstruction task where a partially masked adjacency matrix \textit{A} and a node-feature matrix \textit{X} are used to predict all original edges in the graph. We employ both the VGAE and GAE models discussed in Section \ref{sec:relgae}. In a non-probabilistic setting (GAE) the coreference graph is obtained by passing the adjacency matrix \textit{A} and node-feature matrix \textit{X} through a Graph Convolutional Neural Network (GCN) encoder and then computing the reconstructed matrix $\hat{A}$ from the latent embeddings \textit{Z}:

\[Z= GCN(X,A) \tag{1}\]
\[\hat{A} =\sigma(ZZ^\tau) \tag{2}\]

For a detailed overview of the (probabilistic) variational graph autoencoder we refer the reader to the original paper by \citet{kipf2016variational}.

Our experiments are performed in a cross-document setting, meaning that the input adjacency matrix \textit{A} contains all events in the ENCORE dataset. Following the original approach by \citet{kipf2016variational} we mask 15\% of the edges, 5\% to be used for validation and the remaining 10\% for testing. An equal amount of non-edges is randomly sampled from \textit{A} to balance the validation and test data. 

We extract masked edges and non-edges and use them to build the training, validation and test sets for the mention-pair baseline models detailed above, ensuring that both the mention-pair and graph autoencoder models have access to exactly the same data for training, validation and testing. We define the encoder network with a 64-dimension hidden layer and 32-dimension latent variables. For all experiments we train for a total duration of 200 epochs using an Adam optimizer \cite{kingma2014adam} and a learning rate of 0.001. 

We construct node features through Dutch monolingual transformer models by average-pooling token representations for each token in the event span in the models' final hidden layer, resulting in a 768-dimensional feature vector for each node in the graph. For this we use the Dutch BERTje model \cite{de2019bertje}, a Dutch sentenceBERT model \cite{reimers2019sentence} and the Dutch RoBERTa-based RobBERT model \cite{delobelle2020robbert}. Additionally, we create a second feature set for the BERTje and RobBERT models where each event is represented by the concatenation of the last 4 layers' average-pooled token representations \citet{devlin2018bert}. This in turn results in a 3072-dimensional feature vector. 

Finally, we also evaluate a language-agnostic featureless model where \textit{X} is represented by the identity matrix of \textit{A}. 

\subsubsection{Hardware Specifications}
The baseline coreference algorithms were trained and evaluated on 2 Tesla V100-SXM2-16GB GPUs. Due to GPU memory constraints, the Graph encoder models were all trained and evaluated on a single 2.6 GHz 6-Core Intel Core i7 CPU.  

\section{Results and Discussion}

Results from our experiments are disclosed in Table 1. Results are reported through the CONLL F1 metric, an average of 3 commonly used metrics for coreference evaluation: MUC \cite{vilain1995model}, B$^3$ \cite{bagga1998algorithms} and CEAF \cite{luo2005coreference}. We find that the graph autoencoder models consistently outperform the traditional mention-pair approach. Moreover, we find the autoencoder approach significantly reduces model size, training time and inference speed even when compared to parameter-efficient transformer-based methods. We note that the VGAE models perform slightly worse compared to their non-probabilistic counterparts, which is contrary to the findings in \citet{kipf2016variational}. This can be explained by the use of more complex acyclic graph data in the original paper. In this more uncertain context, probabilistic models would likely perform better. 

\begin{figure}[h]
    \centering
    \includegraphics[width=\columnwidth,height=4cm]{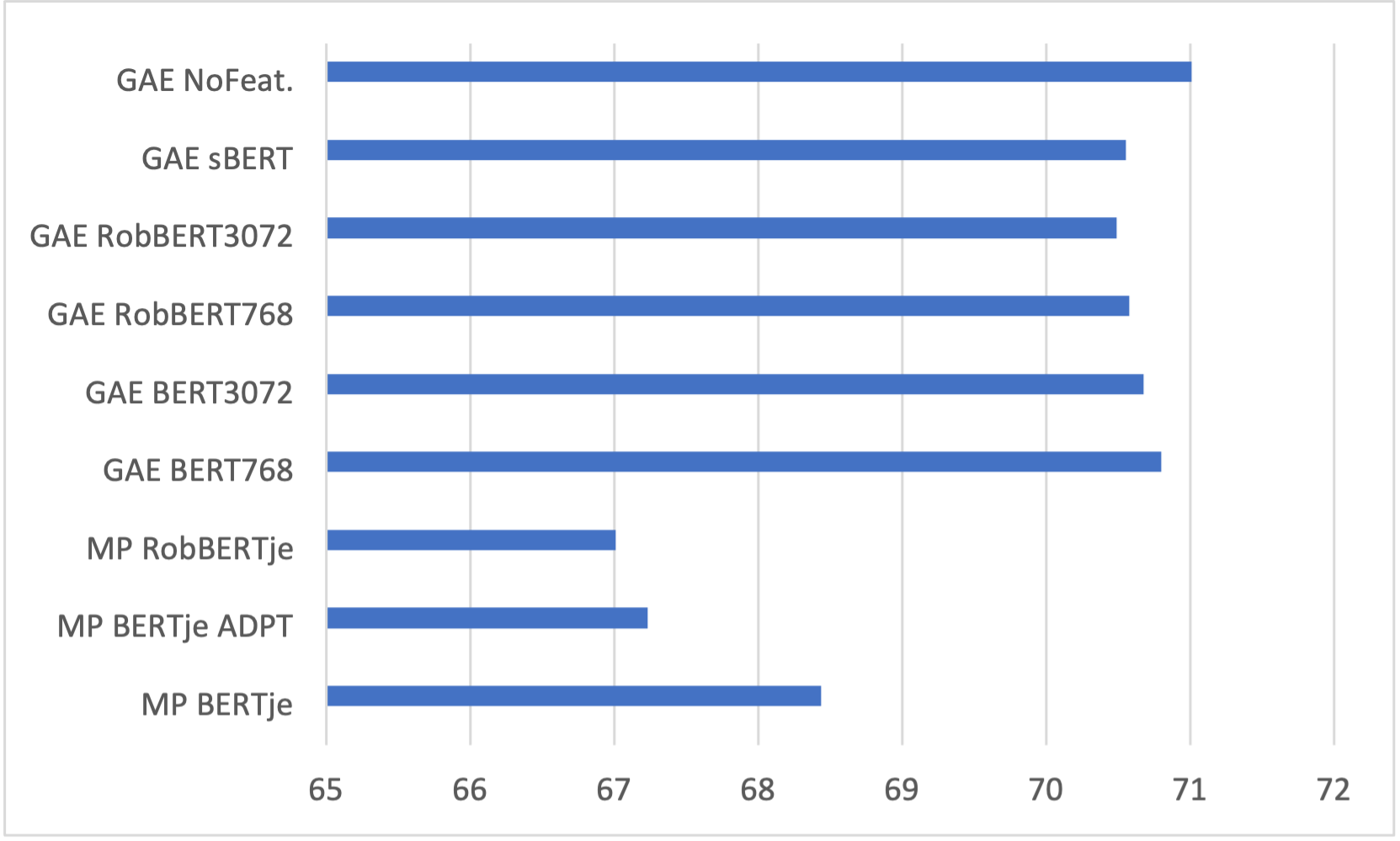}
    \caption{Average Levenshtein distance for True Positive (TP) classifications across all models}
    \label{fig:analysis}
\end{figure}

As a means of quantitative error analysis, we report the average Levenshtein distance between two event spans for the True Positive (TP) pairs in our test set in Figure \ref{fig:analysis}. Logically, if graph-based models are able to better classify harder (i.e non-similar) edges, the average Levenstein distance for predicted TP edges should be higher than for the mention-pair models. For readability's sake we only include results for the best performing GAE-class models. A more detailed table can be found in the Appendix. We find that the average distance between TP pairs increases for our introduced graph models, indicating that graph-based models can, to some extent, mitigate the pitfalls of mention-pair methodologies as discussed in Section \ref{sec:intro}.

\section{Ablation Studies}

\begin{figure}[h]
    \centering
    \includegraphics[width=\columnwidth,height=4cm]{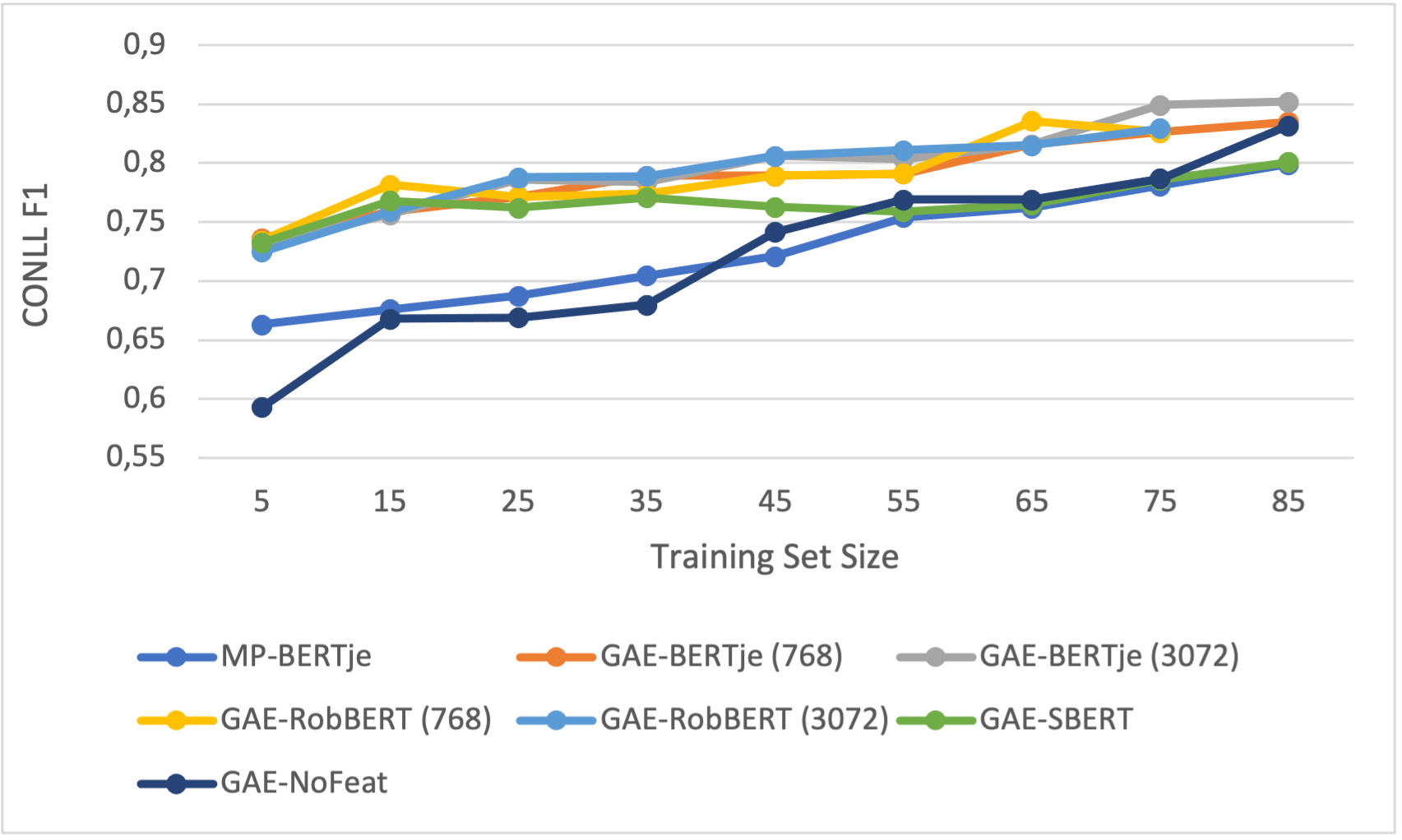}
    \caption{CONLL F1 performance with respect to the available training data.}
    \label{fig:ablation-datasize}
\end{figure}

We gauge the robustness of the graph-based models in low-data settings by re-running the original experiment and continually reducing the available training data by increments of 10\%. Figure \ref{fig:ablation-datasize} shows the CONLL F1 score for each of the models with respect to the available training data size. Also here, only the best-performing GAE-class models are visualized and an overview of all models' performance can be found in the Appendix. Surprisingly, we find that training the model on as little as 5\% of the total amount of edges in the dataset can already lead to satisfactory results. Logically, feature-less models suffer from a significant drop in performance when available training data is reduced. We also find that the overall drop in performance is far greater for the traditional mention-pair model than it is for the feature-based GAE-class models in low-data settings. Overall, we conclude that the introduced family of models can be a lightweight and stable alternative to traditional mention-pair coreference models, even in settings with little to no available training data. 

\section{Conclusion}
We show that ECR through graph autoencoders significantly outperforms traditional mention-pair approaches in terms of performance, speed and model size in settings where coreference chains are at least partially known. Our method provides a fast and lightweight approach for processing large cross-document collections of event data. Additionally, our analysis shows that combining BERT-like embeddings and structural knowledge of coreference chains mitigates the issues in mention-pair classification w.r.t the dependence on surface-form lexical similarity. Our ablation experiments reveal that only a very small number of training edges is needed to obtain satisfactory performance. 

Future work will explore the possibility of combining mention-pair models with the proposed graph autoencoder approach in a pipeline setting in order to make it possible to employ graph reconstruction models in settings where initially all edges in the graph are unknown. Additionally, we aim to perform more fine-grained analyses, both quantitative and qualitative, regarding the type of errors made by graph-based coreference models.  

\section{Limitations}
We identify two possible limitations with the work presented above. First, by framing coreference resolution as a graph reconstruction task we assume that at least some coreference links in the cross-document graph are available to train on. However, we note that this issue can in part be mitigated by a simple exact match heuristic for event spans on unlabeled data. Moreover, in most application settings it is not inconceivable that at least a partial graph is available. 

A second limitation stems from the fact that we modelled coreference chains as undirected graphs. It could be argued that some coreferential relationships such as pronominal anaphora could be more accurately modelled using directed graphs instead.

\bibliography{anthology,custom}

\begin{thebibliography}{39}
\expandafter\ifx\csname natexlab\endcsname\relax\def\natexlab#1{#1}\fi

\bibitem[{Ahmed et~al.(2023)Ahmed, Nath, Martin, and Krishnaswamy}]{ahmed20232}
Shafiuddin~Rehan Ahmed, Abhijnan Nath, James~H Martin, and Nikhil Krishnaswamy. 2023.
\newblock 2n is better than n2: Decomposing event coreference resolution into two tractable problems.
\newblock \emph{arXiv preprint arXiv:2305.05672}.

\bibitem[{Bagga and Baldwin(1998)}]{bagga1998algorithms}
Amit Bagga and Breck Baldwin. 1998.
\newblock Algorithms for scoring coreference chains.
\newblock In \emph{The first international conference on language resources and evaluation workshop on linguistics coreference}, volume~1, pages 563--566. Citeseer.

\bibitem[{Barhom et~al.(2019)Barhom, Shwartz, Eirew, Bugert, Reimers, and Dagan}]{barhom2019revisiting}
Shany Barhom, Vered Shwartz, Alon Eirew, Michael Bugert, Nils Reimers, and Ido Dagan. 2019.
\newblock Revisiting joint modeling of cross-document entity and event coreference resolution.
\newblock \emph{arXiv preprint arXiv:1906.01753}.

\bibitem[{Cattan et~al.(2021{\natexlab{a}})Cattan, Eirew, Stanovsky, Joshi, and Dagan}]{cattan2021cross}
Arie Cattan, Alon Eirew, Gabriel Stanovsky, Mandar Joshi, and Ido Dagan. 2021{\natexlab{a}}.
\newblock Cross-document coreference resolution over predicted mentions.
\newblock \emph{arXiv preprint arXiv:2106.01210}.

\bibitem[{Cattan et~al.(2021{\natexlab{b}})Cattan, Eirew, Stanovsky, Joshi, and Dagan}]{cattan2021realistic}
Arie Cattan, Alon Eirew, Gabriel Stanovsky, Mandar Joshi, and Ido Dagan. 2021{\natexlab{b}}.
\newblock Realistic evaluation principles for cross-document coreference resolution.
\newblock \emph{arXiv preprint arXiv:2106.04192}.

\bibitem[{Chattopadhyay et~al.(2023)Chattopadhyay, Zhang, Wipf, Arora, and Vidal}]{chattopadhyay2023learning}
Aditya Chattopadhyay, Xi~Zhang, David~Paul Wipf, Himanshu Arora, and Ren{\'e} Vidal. 2023.
\newblock Learning graph variational autoencoders with constraints and structured priors for conditional indoor 3d scene generation.
\newblock In \emph{Proceedings of the IEEE/CVF Winter Conference on Applications of Computer Vision}, pages 785--794.

\bibitem[{Chen et~al.(2015)Chen, Xu, Liu, Zeng, and Zhao}]{chen_event_2015}
Yubo Chen, Liheng Xu, Kang Liu, Daojian Zeng, and Jun Zhao. 2015.
\newblock \href {https://doi.org/10.3115/v1/P15-1017} {Event {Extraction} via {Dynamic} {Multi}-{Pooling} {Convolutional} {Neural} {Networks}}.
\newblock \emph{Proceedings of the 53rd Annual Meeting of the Association for Computational Linguistics and the 7th International Joint Conference on Natural Language Processing}, pages 167--176.

\bibitem[{Cybulska and Vossen(2015)}]{cybulska_translating_2015}
Agata Cybulska and Piek Vossen. 2015.
\newblock \href {https://doi.org/10.3115/v1/W15-0801} {Translating {Granularity} of {Event} {Slots} into {Features} for {Event} {Coreference} {Resolution}.}
\newblock In \emph{Proceedings of the {The} 3rd {Workshop} on {EVENTS}: {Definition}, {Detection}, {Coreference}, and {Representation}}, pages 1--10, Denver, Colorado. Association for Computational Linguistics.

\bibitem[{De~Langhe et~al.(2022{\natexlab{a}})De~Langhe, De~Clercq, and Hoste}]{de2022constructing}
Loic De~Langhe, Orph{\'e}e De~Clercq, and Veronique Hoste. 2022{\natexlab{a}}.
\newblock Constructing a cross-document event coreference corpus for dutch.
\newblock \emph{Language Resources and Evaluation}, pages 1--30.

\bibitem[{De~Langhe et~al.(2022{\natexlab{b}})De~Langhe, De~Clercq, and Hoste}]{de2022investigating}
Loic De~Langhe, Orphée De~Clercq, and Veronique Hoste. 2022{\natexlab{b}}.
\newblock Investigating cross-document event coreference for dutch.

\bibitem[{De~Langhe et~al.(2023)De~Langhe, Desot, De~Clercq, and Hoste}]{electronics12040850}
Loic De~Langhe, Thierry Desot, Orphée De~Clercq, and Veronique Hoste. 2023.
\newblock \href {https://doi.org/10.3390/electronics12040850} {A benchmark for dutch end-to-end cross-document event coreference resolution}.
\newblock \emph{Electronics}, 12(4).

\bibitem[{de~Vries et~al.(2019)de~Vries, van Cranenburgh, Bisazza, Caselli, van Noord, and Nissim}]{de2019bertje}
Wietse de~Vries, Andreas van Cranenburgh, Arianna Bisazza, Tommaso Caselli, Gertjan van Noord, and Malvina Nissim. 2019.
\newblock Bertje: A dutch bert model.
\newblock \emph{arXiv preprint arXiv:1912.09582}.

\bibitem[{Delobelle et~al.(2020)Delobelle, Winters, and Berendt}]{delobelle2020robbert}
Pieter Delobelle, Thomas Winters, and Bettina Berendt. 2020.
\newblock Robbert: a dutch roberta-based language model.
\newblock \emph{arXiv preprint arXiv:2001.06286}.

\bibitem[{Delobelle et~al.(2022)Delobelle, Winters, and Berendt}]{delobelle2022robbertje}
Pieter Delobelle, Thomas Winters, and Bettina Berendt. 2022.
\newblock Robbertje: A distilled dutch bert model.
\newblock \emph{arXiv preprint arXiv:2204.13511}.

\bibitem[{Devlin et~al.(2018)Devlin, Chang, Lee, and Toutanova}]{devlin2018bert}
Jacob Devlin, Ming-Wei Chang, Kenton Lee, and Kristina Toutanova. 2018.
\newblock Bert: Pre-training of deep bidirectional transformers for language understanding.
\newblock \emph{arXiv preprint arXiv:1810.04805}.

\bibitem[{Fan et~al.(2022)Fan, Li, Luo, and Xu}]{fan2022enhancing}
Chuang Fan, Jiaming Li, Xuan Luo, and Ruifeng Xu. 2022.
\newblock Enhancing structure preservation in coreference resolution by constrained graph encoding.
\newblock \emph{IEEE/ACM Transactions on Audio, Speech, and Language Processing}, 30:2557--2567.

\bibitem[{Huang et~al.(2022)Huang, Xu, He, Li, and Zhu}]{huang2022incorporating}
Congcheng Huang, Sheng Xu, Longwang He, Peifeng Li, and Qiaoming Zhu. 2022.
\newblock Incorporating generation method and discourse structure to event coreference resolution.
\newblock In \emph{International Conference on Neural Information Processing}, pages 73--84. Springer.

\bibitem[{Humphreys et~al.(1997)Humphreys, Gaizauskas, and Azzam}]{humphreys_event_1997}
Kevin Humphreys, Robert Gaizauskas, and Saliha Azzam. 1997.
\newblock Event coreference for information extraction.
\newblock In \emph{Proceedings of the {ACL}/{EACL} {Workshop} on {Operational} {Factors} in {Practical}, {Robus} {Anaphora} {Resolution} for {Unrestricted} {Texts}}, pages 75--81.

\bibitem[{Joshi et~al.(2020)Joshi, Chen, Liu, Weld, Zettlemoyer, and Levy}]{joshi2020spanbert}
Mandar Joshi, Danqi Chen, Yinhan Liu, Daniel~S Weld, Luke Zettlemoyer, and Omer Levy. 2020.
\newblock Spanbert: Improving pre-training by representing and predicting spans.
\newblock \emph{Transactions of the Association for Computational Linguistics}, 8:64--77.

\bibitem[{Kenyon-Dean et~al.(2018)Kenyon-Dean, Cheung, and Precup}]{kenyon-dean_resolving_2018}
Kian Kenyon-Dean, Jackie Chi~Kit Cheung, and Doina Precup. 2018.
\newblock \href {http://www.aclweb.org/anthology/S18-2001} {Resolving {Event} {Coreference} with {Supervised} {Representation} {Learning} and {Clustering}-{Oriented} {Regularization}}.
\newblock In \emph{Proceedings of the {Seventh} {Joint} {Conference} on {Lexical} and {Computational} {Semantics}}, pages 1--10, New Orleans, Louisiana. Association for Computational Linguistics.

\bibitem[{Kingma and Ba(2014)}]{kingma2014adam}
Diederik~P Kingma and Jimmy Ba. 2014.
\newblock Adam: A method for stochastic optimization.
\newblock \emph{arXiv preprint arXiv:1412.6980}.

\bibitem[{Kipf and Welling(2016{\natexlab{a}})}]{kipf2016semi}
Thomas~N Kipf and Max Welling. 2016{\natexlab{a}}.
\newblock Semi-supervised classification with graph convolutional networks.
\newblock \emph{arXiv preprint arXiv:1609.02907}.

\bibitem[{Kipf and Welling(2016{\natexlab{b}})}]{kipf2016variational}
Thomas~N Kipf and Max Welling. 2016{\natexlab{b}}.
\newblock Variational graph auto-encoders.
\newblock \emph{arXiv preprint arXiv:1611.07308}.

\bibitem[{Li et~al.(2020)Li, Fabbri, Hingmire, and Radev}]{li2020r}
Irene Li, Alexander Fabbri, Swapnil Hingmire, and Dragomir Radev. 2020.
\newblock R-vgae: Relational-variational graph autoencoder for unsupervised prerequisite chain learning.
\newblock \emph{arXiv preprint arXiv:2004.10610}.

\bibitem[{Liu et~al.(2018)Liu, Allamanis, Brockschmidt, and Gaunt}]{liu2018constrained}
Qi~Liu, Miltiadis Allamanis, Marc Brockschmidt, and Alexander Gaunt. 2018.
\newblock Constrained graph variational autoencoders for molecule design.
\newblock \emph{Advances in neural information processing systems}, 31.

\bibitem[{Liu and Lapata(2019)}]{liu2019hierarchical}
Yang Liu and Mirella Lapata. 2019.
\newblock Hierarchical transformers for multi-document summarization.
\newblock \emph{arXiv preprint arXiv:1905.13164}.

\bibitem[{Lu and Ng(2021)}]{lu2021conundrums}
Jing Lu and Vincent Ng. 2021.
\newblock Conundrums in event coreference resolution: Making sense of the state of the art.
\newblock In \emph{Proceedings of the 2021 Conference on Empirical Methods in Natural Language Processing}, pages 1368--1380.

\bibitem[{Luo(2005)}]{luo2005coreference}
Xiaoqiang Luo. 2005.
\newblock On coreference resolution performance metrics.
\newblock In \emph{Proceedings of Human Language Technology Conference and Conference on Empirical Methods in Natural Language Processing}, pages 25--32.

\bibitem[{Minard et~al.(2016)Minard, Speranza, Urizar, van Erp, Schoen, and van Son}]{minard_meantime_2016}
Anne-Lyse Minard, Manuela Speranza, Ruben Urizar, Marieke van Erp, Anneleen Schoen, and Chantal van Son. 2016.
\newblock {MEANTIME}, the {NewsReader} {Multilingual} {Event} and {Time} {Corpus}.
\newblock In \emph{Proceedings of the 10th language resources and evaluation conference ({LREC} 2016)}, page~6, Portorož, Slovenia. European Language Resources Association (ELRA).

\bibitem[{Mitamura et~al.(2015)Mitamura, Yamakawa, Holm, Song, Bies, Kulick, and Strassel}]{mitamura_event_2015}
Teruko Mitamura, Yukari Yamakawa, Susan Holm, Zhiyi Song, Ann Bies, Seth Kulick, and Stephanie Strassel. 2015.
\newblock \href {https://doi.org/10.3115/v1/W15-0809} {Event {Nugget} {Annotation}: {Processes} and {Issues}}.
\newblock In \emph{Proceedings of the {The} 3rd {Workshop} on {EVENTS}: {Definition}, {Detection}, {Coreference}, and {Representation}}, pages 66--76, Denver, Colorado. Association for Computational Linguistics.

\bibitem[{Nguyen et~al.(2016)Nguyen, Meyers, and Grishman}]{nguyen2016new}
Thien~Huu Nguyen, Adam Meyers, and Ralph Grishman. 2016.
\newblock New york university 2016 system for kbp event nugget: A deep learning approach.
\newblock In \emph{TAC}.

\bibitem[{NIST(2005)}]{nist_ace_2005}
NIST. 2005.
\newblock The {ACE} 2005 ( {ACE} 05 ) {Evaluation} {Plan}.

\bibitem[{Pfeiffer et~al.(2020)Pfeiffer, R{\"u}ckl{\'e}, Poth, Kamath, Vuli{\'c}, Ruder, Cho, and Gurevych}]{pfeiffer2020adapterhub}
Jonas Pfeiffer, Andreas R{\"u}ckl{\'e}, Clifton Poth, Aishwarya Kamath, Ivan Vuli{\'c}, Sebastian Ruder, Kyunghyun Cho, and Iryna Gurevych. 2020.
\newblock Adapterhub: A framework for adapting transformers.
\newblock \emph{arXiv preprint arXiv:2007.07779}.

\bibitem[{Reimers and Gurevych(2019)}]{reimers2019sentence}
Nils Reimers and Iryna Gurevych. 2019.
\newblock Sentence-bert: Sentence embeddings using siamese bert-networks.
\newblock \emph{arXiv preprint arXiv:1908.10084}.

\bibitem[{Tran et~al.(2021)Tran, Phung, and Nguyen}]{tran2021exploiting}
Hieu~Minh Tran, Duy Phung, and Thien~Huu Nguyen. 2021.
\newblock Exploiting document structures and cluster consistencies for event coreference resolution.
\newblock In \emph{Proceedings of the 59th Annual Meeting of the Association for Computational Linguistics and the 11th International Joint Conference on Natural Language Processing (Volume 1: Long Papers)}, pages 4840--4850.

\bibitem[{Vermeulen(2018)}]{8612730}
Judith Vermeulen. 2018.
\newblock newsdna : promoting news diversity : an interdisciplinary investigation into algorithmic design, personalization and the public interest (2018-2022).

\bibitem[{Vilain et~al.(1995)Vilain, Burger, Aberdeen, Connolly, and Hirschman}]{vilain1995model}
Marc Vilain, John~D Burger, John Aberdeen, Dennis Connolly, and Lynette Hirschman. 1995.
\newblock A model-theoretic coreference scoring scheme.
\newblock In \emph{Sixth Message Understanding Conference (MUC-6): Proceedings of a Conference Held in Columbia, Maryland, November 6-8, 1995}.

\bibitem[{Wu et~al.(2022)Wu, Huang, Fung, and Ji}]{wu2022cross}
Xueqing Wu, Kung-Hsiang Huang, Yi~Fung, and Heng Ji. 2022.
\newblock Cross-document misinformation detection based on event graph reasoning.
\newblock In \emph{Proceedings of the 2022 Conference of the North American Chapter of the Association for Computational Linguistics: Human Language Technologies}, pages 543--558.

\bibitem[{Yang et~al.(2020)Yang, Zhang, Wang, Li, Kim, Walker, Xiao, and Han}]{yang2020relation}
Carl Yang, Jieyu Zhang, Haonan Wang, Sha Li, Myungwan Kim, Matt Walker, Yiou Xiao, and Jiawei Han. 2020.
\newblock Relation learning on social networks with multi-modal graph edge variational autoencoders.
\newblock In \emph{Proceedings of the 13th International Conference on Web Search and Data Mining}, pages 699--707.

\end{thebibliography}
\bibliographystyle{acl_natbib}
\newpage
\appendix

\section{Appendix}
\label{sec:appendix}

\begin{table}[h]
\scalebox{0.8}{
\begin{tabular}{|c|c|}
\hline
\textbf{Model}                           & \textbf{Levenshtein Distance (TP)} \\ \hline
MP RobBERTje                             &            67.01                        \\ \hline
MP BERTje (ADPT)                         &           67.23                         \\ \hline
MP BERTje                                &           68.44                         \\ \hline
GAE NOFEAT                               &          71.01                          \\ \hline
GAE BERTje (768)                         &            70.8                        \\ \hline
GAE BERTje (3072)                        &           70.68                         \\ \hline
GAE RobBERT (768)                        &          70.57                          \\ \hline
GAE RobBERT (3072)                       &            70.49                        \\ \hline
GAE SBERT                                &             70.55                       \\ \hline
VGAE NOFEAT                              &              69.95                      \\ \hline
VGAE BERTje (768)                        &             68.71                       \\ \hline
VGAE BERTje (3072) &     70.04         \\ \hline
VGAE RobBERT (768)  &     70.21          \\ \hline
VGAE RobBERT (3072) &   70.15           \\ \hline
VGAE SBERT          &     70.04          \\ \hline
\end{tabular}}
\caption{Average Levenshtein distance for each True Positive (TP) pair in the test set indicating how well each model predicts comparatively more difficult coreference links.}
\end{table}

\begin{table}[h]
\scalebox{0.8}{
\begin{tabular}{|c|c|c|c|c|c|c|c|c|}
\hline
\textbf{Model}     & \textbf{5} & \textbf{15} & \textbf{25} & \textbf{35} & \textbf{45} & \textbf{55} & \textbf{65} & \textbf{75} \\ \hline
MP RobBERTje       &     0.627       &     0.631        &      0.667       &     0.683        &    0.701         &     0.736        &      0.753       &    0.766         \\ \hline
MP BERTje (ADPT)   &      0.638      &     0.640        &      0.662       &      0.685       &      0.692       &      0.724       &      0.729       &    0.754         \\ \hline
MP BERTje          &    0.663        &     0.675        &   0.687          &      0.704       &    0.721         &     0.754        &   0.762          &  0.781           \\ \hline
GAE NOFEAT         &     0.593       &   0.667          &     0.669        &     0.679        &        0.747     &      0.769       &    0.769         &  0.786           \\ \hline
GAE BERTje (768)   &      0.736      &    0.759         &    0.771         &     0.789        &      0.789       &    0.791         &     0.815        &    0.826         \\ \hline
GAE BERTje (3072)  &      0.730      &    0.756         &     0.786        &       0.784      &    0.805         &     0.803        &     0.815        &    0.849         \\ \hline
GAE RobBERT (768)  &     0.734       &    0.781         &     0.771        &       0.774      &     0.783        &      0.791       &     0.835        &     0.826        \\ \hline
GAE RobBERT (3072) &        0.725    &      0.759       &       0.788      &     0.788        &      0.806       &     0.810        &     0.815        &     0.829        \\ \hline
GAE SBERT          &    0.732        &     0.768        &       0.762      &       0.770      &      0.762       &     0.759        &     0.765        &     0.786        \\ \hline
VGAE NOFEAT        &     0.632       &    0.653         &    0.742         &     0.752        &     0.747        &    0.766         &     0.781        &     0.786        \\ \hline
VGAE BERTje (768)  &        0.672    &     0.747        &     0.753        &    0.758         &     0.758        &   0.773          &   0.795          &     0.809        \\ \hline
VGAE BERTje (3072) &    0.712        &     0.769        &    0.781         &     0.780        &       0.776      &   0.818          &      0.802       &     0.818        \\ \hline
VGAE RobBERT (768)  &     0.672       &   0.745          &    0.757         &      0.758       &      0.759       &     0.770        &  0.791           &       0.799      \\ \hline
VGAE RobBERT (3072) &      0.691      &     0.753        &    0.762         &      0.764       &     0.761        &      0.791       &     0.800        &       0.801      \\ \hline
VGAE SBERT          &    0.651        &    0.681         &       0.735      &     0.738        &     0.726        &     0.711        &     0.745        &     0.735        \\ \hline
\end{tabular}}
\caption{Results (CONLL F1) for the ablation experiments for each individual model. Columns indicate the percentagewise amount of available training data w.r.t the overall size of the ENCORE dataset.}
\end{table}

\end{document}